\documentclass{article}

\usepackage{microtype}
\usepackage{graphicx}
\usepackage{booktabs}
\usepackage{hyperref}

\newcommand{\SMACINDEFFRANK}{0.29 \pm 0.02}
\newcommand{\SMACINDDAQ}{1.23 \pm 0.06}
\newcommand{\SMACINDPROBE}{0.73 \pm 0.05}
\newcommand{\SMACINDWIN}{0.48 \pm 0.18}
\newcommand{\SMACSHREFFRANK}{0.31 \pm 0.03}
\newcommand{\SMACSHRDAQ}{1.07 \pm 0.20}
\newcommand{\SMACSHRPROBE}{0.75 \pm 0.05}
\newcommand{\SMACSHRWIN}{0.39 \pm 0.07}
\newcommand{\SMACMASKINDEFFRANK}{0.29 \pm 0.01}
\newcommand{\SMACMASKINDDAQ}{1.13 \pm 0.08}
\newcommand{\SMACMASKINDPROBE}{0.49 \pm 0.04}
\newcommand{\SMACMASKINDWIN}{0.39 \pm 0.02}
\newcommand{\SMACMASKSHREFFRANK}{0.30 \pm 0.02}
\newcommand{\SMACMASKSHRDAQ}{0.98 \pm 0.18}
\newcommand{\SMACMASKSHRPROBE}{0.49 \pm 0.14}
\newcommand{\SMACMASKSHRWIN}{0.24 \pm 0.07}

\usepackage[accepted]{icml2026}
\usepackage{amsmath}
\usepackage{amssymb}
\usepackage{mathtools}
\usepackage[capitalize,noabbrev]{cleveref}

\newcommand{\parnobf}[1]{\vspace{0.25em}\noindent\textbf{#1.}}

\graphicspath{{figures/}{samples/figures/}}
\hypersetup{colorlinks=true, linkcolor=blue, urlcolor=blue, citecolor=blue}

\icmltitlerunning{Feedback Attribution and Representation Geometry in MARL}

\begin{document}

\twocolumn[
  \icmltitle{
  Feedback Attribution and Representation Geometry:\\
  Metrics for Comparing Individual and Shared Rewards in MARL}

  \begin{icmlauthorlist}
    \icmlauthor{Tasha Pais}{softmax}
    \icmlauthor{Richard Higgins}{softmax}
  \end{icmlauthorlist}

  \icmlaffiliation{softmax}{Softmax, San Francisco, CA, USA}
  \icmlcorrespondingauthor{Tasha Pais}{tasha@softmax.com}
  \icmlkeywords{multi-agent reinforcement learning, world feedback, credit assignment, representation geometry, evaluation metrics, embedding effective rank}

  \vskip 0.15in
  {\centering \includegraphics[width=0.94\linewidth]{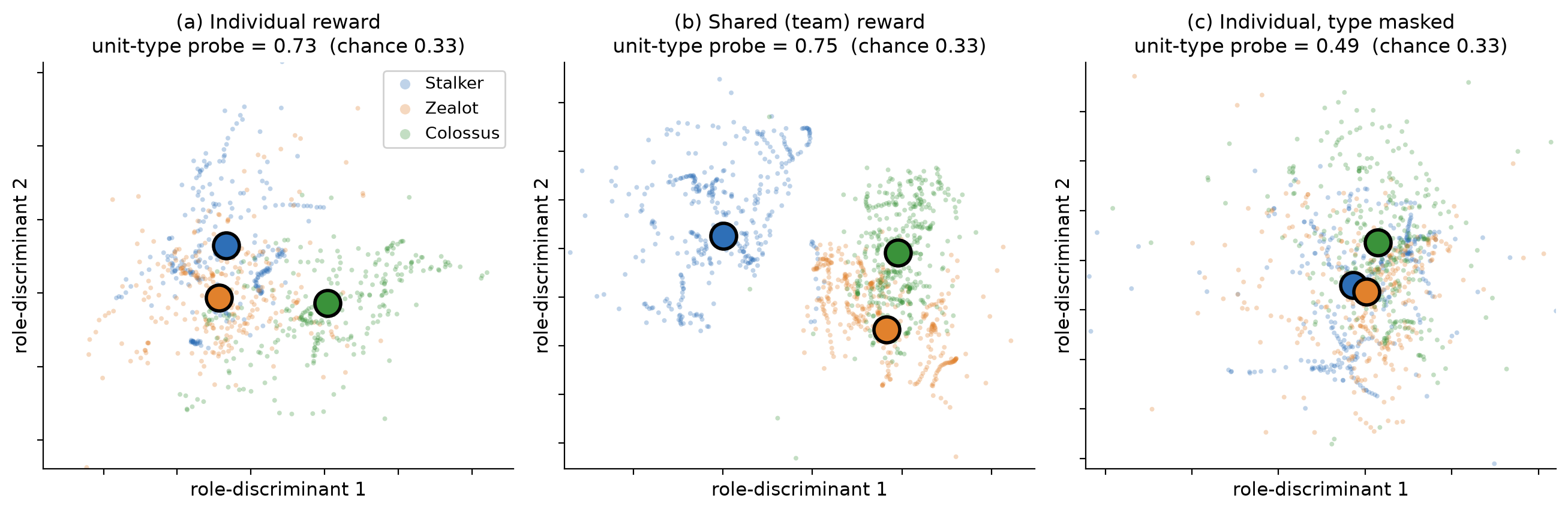} \par}
  \refstepcounter{figure}\label{fig:teaser}
  \vskip 0.03in
  {\small \textbf{Figure~\thefigure: Role-separable geometry tracks observation, not reward attribution.}
  Per-agent GRU embeddings from competent MAPPO policies on SMACv2 \texttt{protoss\_5\_vs\_5} are shown in a linear projection chosen to separate unit types (large circles: role centroids; small dots: per-timestep embeddings). The projection is fit on three of five ally slots and plotted only on the two slots held out during fitting, so role structure must generalize across agents rather than memorize slot identity. (a) With individual reward, the three protoss roles occupy distinct regions (probe $0.73$, chance $1/3$). (b) With shared team reward, the geometry is essentially unchanged (probe $0.75$). (c) Masking each agent's own unit-type bits collapses the clusters and drops the probe to $0.49$. Across the comparison, masking observation changes role geometry, while reward attribution appears mainly in behavior.}
  \vskip 0.2in
]

\printAffiliationsAndNotice{Workshop submission: \emph{RLxF: Reinforcement Learning from World Feedback}.}

\begin{abstract}
Cooperative multi-agent RL systems routinely use team-averaged rewards, a feedback-attribution choice that gives each agent the team outcome regardless of its individual contribution.
We ask whether this leaves a measurable signature, geometric or behavioral, on learned representations.
We propose EffRank/$n$ (effective rank normalized by agent count) and $D_\text{act}$ (mean pairwise KL divergence between agents' action distributions) as low-overhead diagnostics for reward-attribution effects, then test them on competent MAPPO agents in SMACv2 \texttt{protoss\_5\_vs\_5}, where unit type is encoded in the observation.
In an observation $\times$ reward-attribution comparison (unit type observed vs.\ masked; individual damage-contribution reward vs.\ shared team reward), geometry follows observation rather than reward.
With unit type observed, shared and individual rewards have similar EffRank/$n$ ($0.31{\pm}0.03$ vs.\ $0.29{\pm}0.02$) and probe accuracy ($0.75{\pm}0.05$ vs.\ $0.73{\pm}0.05$, both $\gg 1/3$ chance), while $D_\text{act}$ leans higher under individual rewards ($1.23{\pm}0.06$ vs.\ $1.07{\pm}0.20$).
Masking unit type cuts the above-chance probe signal by more than half, to $0.49$ in both reward arms.
In short: individually rewarded agents are competent and separable by role, but on SMACv2 the observation explains the geometry and reward attribution shows up mainly in behavior.
Thus geometric diagnostics must control for observed role information and test persistent roles that are not directly observed.
EffRank/$n$ and $D_\text{act}$ add $<$5\% overhead.
\end{abstract}

\section{Introduction}
\label{sec:intro}

Multi-agent reinforcement learning turns reward design into a feedback-attribution problem.
In single-agent RL, the attribution question is trivial: the agent receives the reward for its own action.
In cooperative tasks, practitioners often replace per-agent contribution with a single team-averaged reward \citep{yu2022mappo,rashid2020qmix}.
That choice aligns agents with a joint objective, but it also gives every agent the same feedback regardless of which action caused the outcome.
The credit-assignment literature treats this as an optimization problem \citep{foerster2018counterfactual}; here we ask whether individual vs.\ shared rewards leave a measurable trace in the shared representation or behavior.
We give diagnostics that separate two effects: reward attribution can change behavior, while directly observed roles can explain apparent role geometry.

The motivating failure mode is familiar in large-scale MARL.
Neural MMO \citep{suarez2019neuralmmo} reports homogeneous strategies under team rewards where complementary roles would be useful, especially at 100 to 1000 agents.
A natural geometric hypothesis is that shared feedback compresses the shared encoder below the dimensionality needed for role differentiation.
We introduce diagnostics for that hypothesis, then use SMACv2 to show the main confound: when role information is in the observation, a high role probe can reflect observed role input rather than specialization produced by the reward.
The right comparison is therefore reward attribution after controlling for what the observation already supplies.

\parnobf{Contributions}
First, we define two minibatch-computable diagnostics: EffRank/$n$ for representation geometry and $D_\text{act}$ for action-distribution diversity.
Second, we validate them on competent SMACv2 \texttt{protoss\_5\_vs\_5} policies trained for 4M environment steps.
Third, we run an observation $\times$ reward-attribution comparison and find that unit-type geometry follows observation, not reward, while reward attribution still affects behavior.
Finally, we apply the same diagnostics to a Tribal Village environment with 48 agents.

\section{Related Work}
\label{sec:related}

\parnobf{Large-scale MARL}
Neural MMO \citep{suarez2019neuralmmo,suarez2023neuralmmo} studies large agent populations in persistent, open-ended worlds and reports that shared team rewards can produce homogeneous strategies instead of complementary roles.
The 2.0 benchmark expands the setting to many tasks and agents, making role formation and specialization visible at population scale.
Competition results \citep{suarez2024challenge} show that strong agents often use individual reward shaping to break symmetry in these large-team settings.
This motivates checking whether shared encoders actually retain role information under different reward attributions.

\parnobf{Credit assignment}
COMA \citep{foerster2018counterfactual} uses a counterfactual baseline to assign credit to individual agents under a shared return.
VDN \citep{sunehag2018vdn} decomposes a team value into per-agent value terms, while QMIX \citep{rashid2020qmix} learns a monotonic mixing network for centralized training and decentralized execution.
These methods keep the cooperative team objective while changing how value or policy updates are attributed to each agent.
Our setting keeps the learning algorithm fixed and measures the representation and behavior that result from different reward attributions.

\parnobf{Representation collapse}
Rank-collapse work in single-agent RL \citep{kumar2021dr3,dormant2023} shows that RL training can reduce representation dimensionality and leave networks with inactive or redundant features.
PRISM \citep{kim2025prism} uses spectral parameter sharing for multi-agent policies, and LoRASA \citep{lorasa2025} uses low-rank agent-specific adaptation for multi-agent policy learning.
Both multi-agent methods add architectural capacity for agent-specific behavior while retaining parameter sharing.
Our metrics ask whether that kind of agent or role separation is present in a trained shared encoder.

\parnobf{Behavioral diversity and roles}
SND \citep{bettini2023snd} measures behavioral heterogeneity in multi-agent learning.
Role diversity \citep{hu2022rolediversity} diagnoses emergent roles, while ACORM \citep{acorm2023} and MA2CL \citep{ma2cl2023} use contrastive objectives to learn or encourage role representations.
Together, this line of work treats role differentiation as a behavioral or representation-learning target rather than only a scalar return objective.
We use the same concern diagnostically, measuring behavior and representation geometry after training.

\section{Method}
\label{sec:method}
\label{sec:metrics}

Let $z_t^{(i)} \in \mathbb{R}^d$ be agent $i$'s shared-encoder embedding at timestep $t$; stack minibatch embeddings in $Z$.

\parnobf{EffRank/$n$}
The effective rank \citep{roy2007effective} of $Z$ estimates embedding dimensionality:
\begin{equation}
  \text{EffRank}(Z)=\exp\!\left(-\sum_i \bar{\sigma}_i \log \bar{\sigma}_i\right),
  \qquad
  \bar{\sigma}_i=\frac{\sigma_i}{\sum_j \sigma_j},
\end{equation}
where $\sigma_i$ are singular values.
In MARL with $n$ agents and $k$ role types, a shared encoder that separates roles need only maintain $k$ geometrically distinct directions.
Normalizing by $n$ gives EffRank/$n$: the expected ceiling for a clean role representation is $k/n$ (one direction per role), while values near $1/n$ indicate collapse to a single shared direction.
We compute it on per-agent time-averaged GRU-embedding centroids so the metric reflects agent/role geometry rather than every transient timestep.
In practice we read EffRank/$n$ comparatively across matched conditions, together with the probe.
The intended interpretation is diagnostic: reward signals that preserve per-agent gradient information should make it easier for the shared encoder to maintain distinct role directions, while averaging those signals can remove the pressure to separate them.

\parnobf{$D_\text{act}$}
We measure behavioral diversity with mean pairwise KL divergence between agents' action distributions:
\begin{equation}
  D_\text{act}=\frac{1}{\binom{n}{2}}\sum_{i \neq j}
  D_\text{KL}\!\left(\pi(\cdot \mid o_i)\,\|\,\pi(\cdot \mid o_j)\right).
\end{equation}
$D_\text{act}$ is an instantaneous marginal, so it captures one-step action divergence rather than role structure that only unfolds over a trajectory.
Because the policy is shared, its absolute value also reflects observation diversity as much as learned specialization.
We therefore compare it across matched reward conditions that hold the environment and observation distribution fixed, and read it with EffRank/$n$ and the probe rather than as a standalone behavioral test.
In this paper, $D_\text{act}$ is informative when it moves while EffRank/$n$ and the probe do not: that pattern says reward attribution changed behavior without changing the decoded role geometry.

\parnobf{Role probe}
We freeze the encoder and fit a logistic regression for the three-way unit-type classification (stalker, zealot, colossus) with leave-one-agent-slot-out validation.
Accuracy above chance ($1/3$ in SMACv2 \texttt{protoss\_5\_vs\_5}) checks whether geometric diversity corresponds to decodable role information; chance accuracy indicates collapse of this particular role label.
Above-chance accuracy still has to be read against observation: if role identity was observed, decoding it does not prove a reward-attribution effect.

\begin{figure}[t]
\centering
\includegraphics[width=0.49\linewidth]{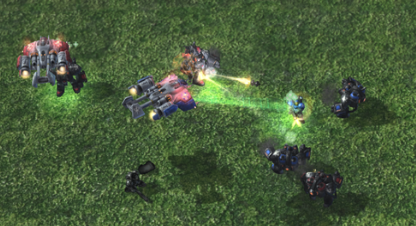}
\includegraphics[width=0.49\linewidth]{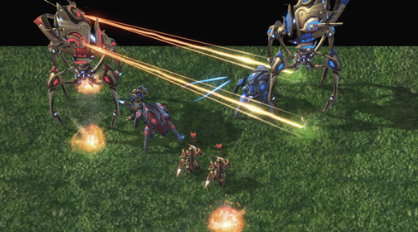}
\caption{\textbf{SMACv2 \texttt{protoss\_5\_vs\_5} (\texttt{10gen\_protoss}).}
Five allied protoss units face five enemies; each ally is a stalker, zealot, or colossus sampled i.i.d.\ per episode. Unit type is observed by default and masked in the ablation, letting us test whether geometry comes from reward attribution or directly observed role identity. Agents are trained for 4M steps ($\sim$24 to 48\% deterministic evaluation win rate across conditions).}
\label{fig:smacv2}
\end{figure}

\begin{table}[t]
\caption{\textbf{Unit type observed.} 4M steps, mean $\pm$ std over 3 seeds and 32 deterministic evaluation episodes per seed. Individual reward gives each agent reward proportional to its damage contribution; shared reward gives every agent the same team-averaged reward. Shared reward does not collapse EffRank/$n$ or the unit-type probe; only behavioral $D_\text{act}$ is higher for individual.}
\label{tab:main}
\resizebox{\columnwidth}{!}{%
\begin{tabular}{lrrr}
\toprule
Feedback & EffRank/$n$ & $D_\text{act}$ & Probe \\
\midrule
Individual & $\SMACINDEFFRANK{}$ & $\mathbf{\SMACINDDAQ{}}$ & $\SMACINDPROBE{}$ \\
Shared     & $\SMACSHREFFRANK{}$ & $\SMACSHRDAQ{}$ & $\SMACSHRPROBE{}$ \\
\bottomrule
\end{tabular}}
\end{table}

\begin{table}[t]
\caption{\textbf{Observation $\times$ reward-attribution comparison.} The four SMACv2 conditions compare unit type observed vs.\ masked with individual damage-contribution reward vs.\ shared team reward, all at 4M steps. EffRank/$n$ is centroid effective rank; $D_\text{act}$ is pairwise KL computed only over actions available to both agents; Probe decodes unit type (chance $=1/3$); Win is deterministic evaluation win rate. Masking unit type cuts the probe from $\sim$0.74 to $0.49$ in both reward arms, while individual rewards mainly improve behavior ($D_\text{act}$ and win rate).}
\label{tab:smacv2}
\resizebox{\columnwidth}{!}{%
\begin{tabular}{llrrrr}
\toprule
Unit type & Feedback & EffRank/$n$ & $D_\text{act}$ & Probe & Win \\
\midrule
Observed & Individual & $\SMACINDEFFRANK{}$ & $\SMACINDDAQ{}$ & $\SMACINDPROBE{}$ & $\SMACINDWIN{}$ \\
Observed & Shared     & $\SMACSHREFFRANK{}$ & $\SMACSHRDAQ{}$ & $\SMACSHRPROBE{}$ & $\SMACSHRWIN{}$ \\
Masked   & Individual & $\SMACMASKINDEFFRANK{}$ & $\SMACMASKINDDAQ{}$ & $\SMACMASKINDPROBE{}$ & $\SMACMASKINDWIN{}$ \\
Masked   & Shared     & $\SMACMASKSHREFFRANK{}$ & $\SMACMASKSHRDAQ{}$ & $\SMACMASKSHRPROBE{}$ & $\SMACMASKSHRWIN{}$ \\
\bottomrule
\end{tabular}}
\end{table}

\parnobf{SMACv2 Environment}
We use SMACv2 \citep{ellis2022smacv2} \texttt{protoss\_5\_vs\_5} (\texttt{10gen\_protoss}): five allied units fight five enemies, and each allied unit type is drawn i.i.d.\ per episode from stalker, zealot, and colossus.
Unit type is the role label.
By default it is encoded directly in each agent's observation through unit-type bits, so SMACv2 can test whether role geometry is coming from visible role input.
Masked runs zero each agent's own unit-type bits, but keep ally and enemy state features.
Because unit types are drawn i.i.d.\ by slot, masking an agent's own bits does not reveal type.

\parnobf{SMACv2 Policy}
All agents execute one fully shared MAPPO \citep{yu2022mappo} policy: a recurrent GRU encoder maps each agent observation to an embedding, followed by linear actor and critic heads.
There are no agent-specific parameters or role-conditioned modules, so any role structure must appear inside the shared encoder.

\parnobf{Comparison}
We train each condition for 4M steps and evaluate 32 deterministic episodes per seed, taking the most likely valid action at each step.
The comparison crosses reward attribution with observation.
Reward is either individual, proportional to each agent's damage contribution, or shared, giving every agent the same team-averaged reward.
Observation is either default unit type observed or masked via \texttt{obs\_mask\_unit\_type}, which zeros each agent's own unit-type bits while leaving other features intact.
If the conditions with unit type observed are explained by direct unit-type input, masking should pull the probe down; we then ask whether individual rewards recover structure without that input, with 3 seeds per cell.

\parnobf{Metric computation}
At 4M steps all four policy families are above the near-zero random baseline against the built-in AI, with deterministic evaluation win rates from $0.24{\pm}0.07$ to $0.48{\pm}0.18$.
Metrics are computed on deterministic evaluation trajectories: EffRank/$n$ over time-averaged per-agent GRU centroids, $D_\text{act}$ over actions available to both agents, and the leave-one-agent-slot-out unit-type probe.
The leave-one-slot-out split is important: the probe must generalize unit-type information to agent slots not used for fitting, so high accuracy is not just in-slot memorization.
The individual condition conserves the team-level signal but divides reward among agents according to damage contribution.

\parnobf{Tribal Village setup}
\label{sec:tribal-method}
We also train individual- and shared-reward recurrent policies for a multi-agent environment called Tribal Village and compute EffRank/$n$ and $D_\text{act}$.

\section{Experiments}
\label{sec:experiments}

We run three checks on the shared encoder.
First, we confirm that individual rewards produce a competent policy with decodable role geometry.
Second, we switch only the reward to shared team reward and ask whether that geometry disappears.
Third, we mask unit type to test whether the role probe was reading observed role input rather than reward-induced specialization.

\parnobf{Individual reward baseline}
We train individual-reward MAPPO with unit type observed and measure win rate, EffRank/$n$, $D_\text{act}$, and the unit-type probe.
This establishes the competence and role-geometry baseline that a shared-reward collapse could remove.

\parnobf{Shared reward comparison}
We switch only the reward to shared team reward and re-measure the same diagnostics.
If reward attribution drives the observed role geometry, shared reward should erase or weaken it while architecture and observation stay fixed.

\parnobf{Unit-type masking}
We zero each agent's own unit-type bits in both reward arms.
If direct observation drives the geometry, both reward arms should lose role information under masking, and individual rewards should not automatically recover it from reward alone.

\section{Results}
\label{sec:results}

\parnobf{Individual reward separates roles}
With unit type visible, individually rewarded agents reach $0.48{\pm}0.18$ deterministic evaluation win rate against the built-in AI, far above the $\sim$0\% random baseline.
Qualitative results include coordinated focus-fire, and the unit-type probe reaches $0.73{\pm}0.05$, above the $1/3$ chance level.
The corresponding EffRank/$n$ is $0.29{\pm}0.02$, with per-role embedding centroids occupying distinct directions.

\parnobf{Shared reward preserves geometry}
Contrary to a pure feedback-attribution account, switching from individual to shared reward leaves the geometric metrics nearly unchanged: EffRank/$n$ is $0.31{\pm}0.03$ shared vs.\ $0.29{\pm}0.02$ individual, and probe accuracy is $0.75{\pm}0.05$ vs.\ $0.73{\pm}0.05$, both far above chance.
The behavioral metric moves in the expected direction, with $D_\text{act}$ higher under individual rewards ($1.23{\pm}0.06$ vs.\ $1.07{\pm}0.20$), but the geometry does not collapse.
Because both policies observe unit type directly, both can carve the embedding space by unit type even when the reward is shared.

\parnobf{Masking weakens role signal}
Removing each agent's own unit-type bits cuts the probe signal by more than half: shared drops from $0.75$ to $0.49$, and individual drops from $0.73$ to $0.49$, only $\sim$0.16 above chance.
The masked probes remain slightly above chance, plausibly because unit type leaks through dynamics such as speed and attack range, but the three clusters merge toward one another.
Individual rewards improve masked behavior relative to shared rewards ($0.39{\pm}0.02$ vs.\ $0.24{\pm}0.07$ win rate, and $1.13{\pm}0.08$ vs.\ $0.98{\pm}0.18$ $D_\text{act}$) without recovering unit-type geometry: both reward arms have probe accuracy $0.49$, and EffRank/$n$ stays between $0.29$ and $0.31$.
On SMACv2, reward attribution affects behavior more clearly than role-decoding geometry once direct role observation is removed.

\parnobf{Competent Tribal Village agents}
The individual checkpoint selected for highest task return reaches $1102.7/1494.7=73.8\%$ of scripted return, while the shared reward checkpoint reaches $898.2/1536.8=58.4\%$; both win $84.4\%$ of 64 deterministic evaluations.
The learned embeddings reveal contribution group: classifiers distinguishing high- from low-contribution agents score $0.820{\pm}0.049$ for individual reward and $0.873{\pm}0.067$ for shared team reward.
EffRank/$n$ is identical ($0.175$ in both arms) and $D_\text{act}$ is comparable ($3.83$ individual, $4.11$ shared team reward), so Tribal shows that the metrics can be collected on competent larger policies but does not support a reward-arm geometry difference.

\section{Discussion}
\label{sec:discussion}

Across SMACv2, individual vs.\ shared rewards changed behavior more clearly than representation geometry.
Individual rewards raised $D_\text{act}$ and win rate, but with unit type visible both reward arms produced nearly the same EffRank/$n$ and probe accuracy.
Masking unit type, not changing reward, collapsed the unit-type probe from $\sim$0.74 to $0.49$ and merged the role clusters.
Thus on SMACv2, observed role input explains the measured role geometry, while reward attribution appears mainly in action diversity and competence.

In Tribal Village, both competent reward arms encode contribution group, while aggregate EffRank/$n$ and $D_\text{act}$ do not separate the reward arms cleanly.
Together, EffRank/$n$ and probes identify role or contribution information in the encoder, not whether reward attribution created it; $D_\text{act}$ remains the behavioral complement.
Future work should use persistent roles hidden from observation, where individual vs.\ shared rewards can be tested without the observed-role shortcut.

\bibliography{references}
\bibliographystyle{icml2026}

\clearpage
\onecolumn
{\noindent\Large\bfseries Appendix\par}
\appendix
\section{Tribal Village Environment}
\label{app:tribal-village}

\begin{figure}[h]
\centering
\includegraphics[width=\textwidth]{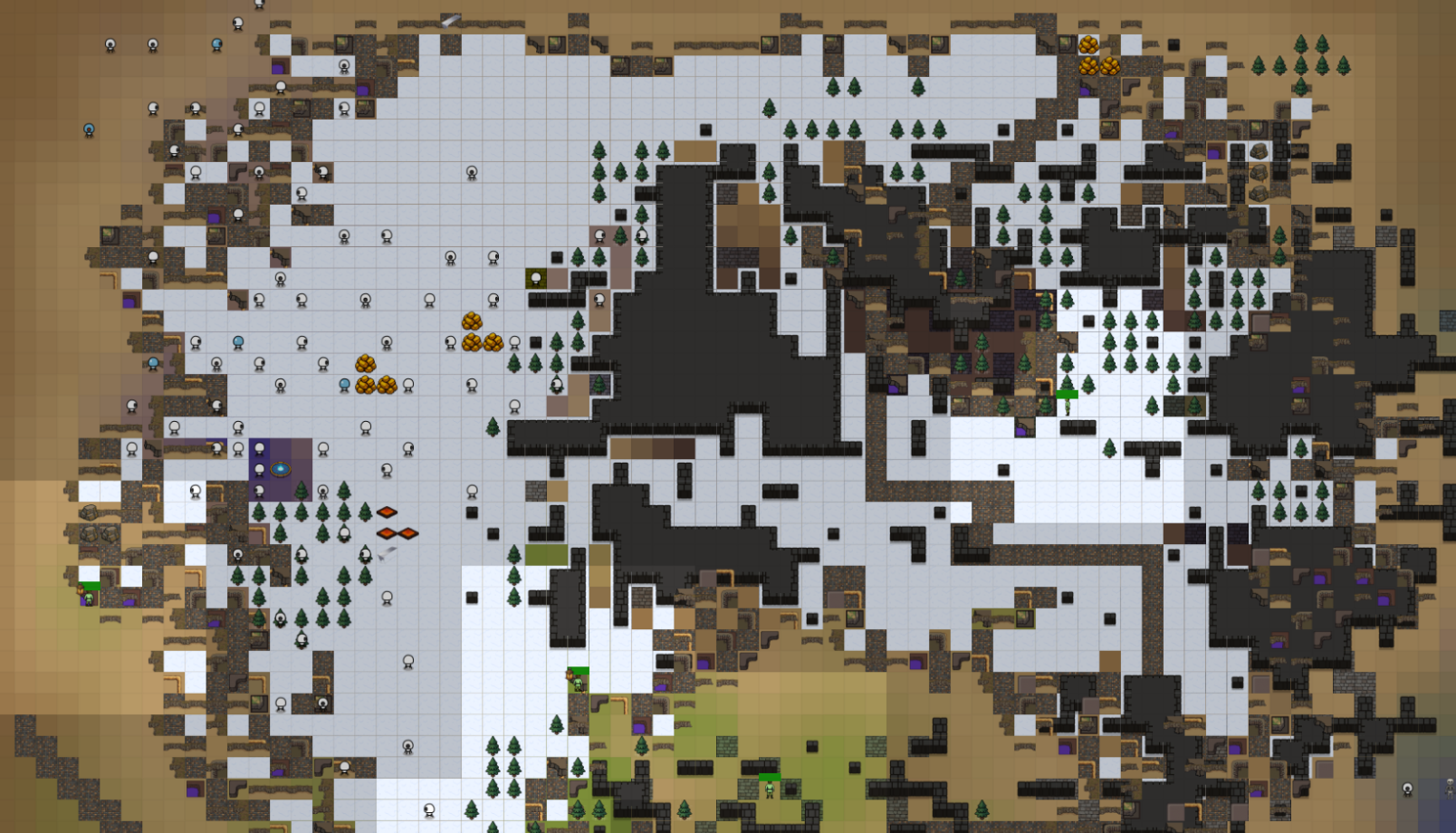}
\caption{Top-down render of the Tribal Village environment used in our
larger-scale evaluation. The map is a grid world in which 48 agents
(the small humanoid sprites) share a single recurrent policy and act
simultaneously. The terrain mixes open traversable tiles (light gray),
impassable rock and walls (dark gray and stone-textured tiles), forest
(pine trees), and bordering desert. Interactable elements include
gatherable gold/ore deposits (yellow clusters), a teleporter tile (the
purple pad, lower left), and hazard tiles (orange diamonds). Agents
forage, build, and contest resources across this shared space, which
yields the 48-agent policies whose EffRank/$n$ and $D_\text{act}$ we
report in Section~\ref{sec:results}.}
\label{fig:tribal-village}
\end{figure}

\end{document}